\documentclass[lettersize,journal]{IEEEtran}
\ifCLASSINFOpdf
\else
\fi

\usepackage{amsmath,amsfonts}
\usepackage{array}
\usepackage{hyperref}
\usepackage{booktabs}
\usepackage{textcomp}
\usepackage{stfloats}
\usepackage{url}
\usepackage{verbatim}
\usepackage{graphicx}
\usepackage{cite}
\usepackage{bbm}
\usepackage{xcolor}
\usepackage{stfloats}
\usepackage{subfigure}
\usepackage{multirow}
\usepackage{enumerate}
\usepackage{colortbl}
\usepackage{threeparttable}
\usepackage[linesnumbered,ruled,vlined]{algorithm2e}

\SetCommentSty{mycommfont}
\SetKwInput{KwInput}{Input}
\SetKwInput{KwOutput}{Output}
\SetKwInput{KwData}{Require}

\begin{document}

\title{Spatial-Temporal Federated Learning for Lifelong Person Re-identification on Distributed Edges}

\author{
Lei Zhang, 
Guanyu Gao,
and Huaizheng Zhang


\thanks{
L. Zhang and G.Y. Gao are with School of Computer Science and Engineering, Nanjing University of Science and Technology, Nanjing 210094, China. Email: \{lei.zhang, gygao\}@njust.edu.cn.
(Corresponding~author:~Guanyu~Gao).

H.Z. Zhang is with School of Computer Science and Engineering, Nanyang Technological University, Singapore. Email: huaizhen001@e.ntu.edu.sg. 

Copyright © 20xx IEEE. Personal use of this material is permitted. However, permission to use this material for any other purposes must be obtained from the IEEE by sending an email to pubs-permissions@ieee.org.
}
}

%



\maketitle

\begin{abstract}
Data drift is a thorny challenge when deploying person re-identification  (ReID) models into real-world devices, where the data distribution is significantly different from that of the training environment and keeps changing.
To tackle this issue, we propose a federated spatial-temporal incremental learning approach, named FedSTIL, which leverages both lifelong learning and federated learning to continuously optimize models deployed on many distributed edge clients. 
Unlike previous efforts, FedSTIL aims to mine spatial-temporal correlations among the knowledge learnt from different edge clients.
Specifically, the edge clients first periodically extract general representations of drifted data to optimize their local models. 
Then, the learnt knowledge from edge clients will be aggregated by centralized parameter server, where the knowledge will be selectively and attentively distilled from spatial- and temporal-dimension with carefully designed mechanisms. 
Finally, the distilled informative spatial-temporal knowledge will be sent back to correlated edge clients to further improve the recognition accuracy of each edge client with a lifelong learning method.
Extensive experiments on a mixture of five real-world datasets demonstrate that our method outperforms others by nearly 4\% in Rank-1 accuracy, while reducing communication cost by 62\%.
All implementation codes are publicly available on \href{https://github.com/MSNLAB/Federated-Lifelong-Person-ReID}{https://github.com/MSNLAB/Federated-Lifelong-Person-ReID}.
\end{abstract}


\section{Introduction}
\IEEEPARstart{P}{erson} re-identification (ReID) aims to retrieve people appearing at different locations and moments from the over-lapped cameras. 
The deep learning-based approaches for person ReID can achieve promising performance on popular benchmarks \cite{shu2021large, zhao2020similarity}, which enables the applications of person ReID in many computer vision-based applications, such as urban analysis, suspect tracking, and city surveillance.

The deployment of person ReID in real-life still suffers from many great challenges.
One prevalent challenge is that the recognition accuracy of the person ReID models will decrease, with the changing of camera environments.
This is mainly because of the domain mismatch between the training and deployment environments.
Specifically, the person ReID models are usually pre-trained on given datasets which consist of images of a fixed set of person identities captured in specific camera environments.
However, in the real-life person ReID deployment, thousands of newly images that are captured every moment often involve many new person identities, which are unavailable at the model training stage. 
Meanwhile, the camera environments are dynamic and ever-changing due to influences brought by many reasons, such as illumination changing and varying camera views.
The domain gap between the training and inference environment limits the performance of person ReID in real-world deployment \cite{pu2021lifelong}
Some preliminary studies \cite{ge2020self, wei2018person} also simulated data drift by training person ReID model on Market-1501 dataset \cite{zheng2015scalable} and testing on MSMT17 dataset \cite{wei2018person}.
These studies observed that model with 73.15\% training mAP can only achieve 4.63\% test mAP, a reduction of almost 68\% due to the domain changes.
%
%

Another challenge for person ReID is to preserve the privacy of the person images \cite{shu2021large}.
The person images contain sensitive private information, such as individuals' identities, locations, genders, ethnicity, and even facial features \cite{zhuang2020performance}.
Sharing these sensitive person images for model training and data analytics is infeasible due to the potential risk of privacy leakage and the expensive communication costs for data transmission. 
Besides, many EU/UK countries have issued privacy protection regulations (e.g., GDPR \cite{EUgdpr}) to prohibit the centralization of sensitive data from in-situ devices.

\begin{figure}[t]
  \centering
  \includegraphics[width=\linewidth]{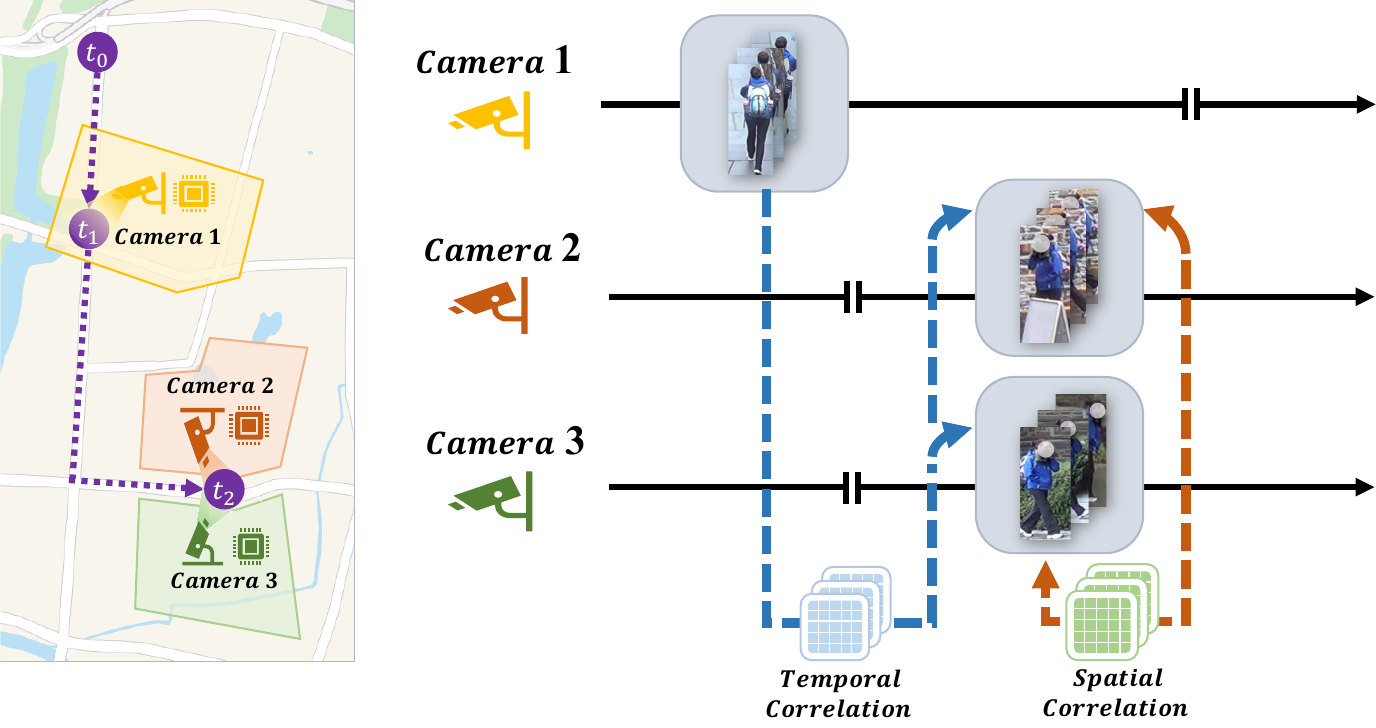}
  \caption{
  The spatial-temporal correlations among the data of different edge clients. 
  The image data captured by different cameras have spatial-temporal correlations, and the edge clients can utilize the spatial-temporal knowledge from others for federated learning to continuously improve their performances.
  }
  \label{fig:background}
\end{figure}

To address the domain drift for person ReID, some recent works (e.g., \cite{pu2021lifelong, wang2016human, zhao2021continual, wu2021generalising, ge2022lifelong}) adopted lifelong learning.
These works enable person ReID models to continuously learn knowledge from new scenarios without forgetting previously learnt knowledge.
HVIL \cite{wang2016human} present a human-in-the-loop paradigm for lifelong person ReID under interactive manual feedback.
AKA \cite{pu2021lifelong} and PTKP\cite{ge2022lifelong} generalised the representation of lifelong person ReID for intra- and inter- domains.
GwFReID \cite{wu2021generalising} alleviated forgetting under a class-imbalance lifelong condition for person ReID.
However, these works require centralized training with the drifted data from deployed devices to learn new knowledge, which also bring data privacy concerns.

To alleviate privacy concerns, some recent works (e.g., \cite{zhuang2020performance, zhang2021edge, sun2021decentralised, yang2022federated}) adopted federated learning to jointly train models on the edge clients.
The federated learning-based approaches enable the sensitive data to be utilized in situ \cite{zhang2021edge, sun2021decentralised, yang2022federated},
while different edge clients can collaboratively update models by aggregating their gradients or parameters \cite{zhuang2020performance}.
FedDG \cite{yang2022federated} generalized person ReID models to tackle cross-edge domain mismatches in federated person ReID.
FedPav \cite{zhuang2020performance} and some other works \cite{sun2021decentralised, zhang2021edge} adopted model aggregation and distillation for the \textit{non-i.i.d} federated person ReID. 
FedReID \cite{wu2021decentralised} proposed an iterative client-server collaborative learning to generalize person ReID models.
SKA \cite{sun2021decentralised} proposed a selective knowledge aggregation method to transfer personalized knowledge among different edge clients.

Prior works, however, considered the problems of continuously updating models and decentralized training models separately.
They are still unable to support distributed edge clients to continuously learn new knowledge while collaboratively sharing their knowledge under privacy-preserving.
Hence, we first propose to judiciously combine both federated learning and lifelong learning for person ReID.
Moreover, we observe that knowledge learnt from different locations and moments has implicit spatial and temporal correlations \cite{shu2021large}.
As illustrated in Fig. \ref{fig:background}, pedestrians that appeared in the past often reappear on other streets in the near future. 
We suppose the knowledge learnt from one edge client may also be informative to other neighbor clients shortly.
However, previous works neglected the spatial-temporal correlations for recognition knowledge at different locations and moments.
Therefore, they failed to adaptively utilize the knowledge across spatial and temporal spaces, and thus limit performance improvement. 

We propose a \textbf{Fed}erated \textbf{S}patial-\textbf{T}emporal \textbf{I}ncremental \textbf{L}earning framework, named FedSTIL, based on spatial-temporal knowledge integration for decentralized continuously learning for person ReID. 
The edge clients utilize their arriving drift data to optimize the local model for incremental domain knowledge.
Meanwhile, some general representations of the drift data are periodically stored in local memory for future rehearsal to alleviate the catastrophic forgetting.
Next, the parameter server integrates the incremental knowledge from different edge clients based on the spatial-temporal correlation.
Then, the parameter server delivers these informative knowledge to the edge clients.
Finally, the edge clients will utilize both the integrated knowledge and previously learnt knowledge to further improve the model with lifelong learning.
%

The main contributions of our paper are summarized as:
\begin{itemize}
\item Propose a federated lifelong person ReID framework, which enables the distributed edge clients to continuously learn incremental knowledge with collaboration. 
\item Design a spatial-temporal knowledge integration method to transfer task-specific knowledge among edge models to improve their performance.
\item Demonstrate the effectiveness of our framework via extensive experiments, ablation studies, and visualization.
\item Release an open-source tool to facilitate the research for federated lifelong person ReID.
\end{itemize}

The rests of this paper are organized as follows. 
Section \ref{sec:related-work} introduces the related works, 
Section \ref{sec:problem-definition} presents the problem definition and the system overview,
Section \ref{sec:learning} illustrates the learning methodology,
Section \ref{sec:experiment} evaluates the performances of our method, 
Section \ref{sec:future_work} discusses some future directions,
and Section \ref{sec:conclusion} concludes the paper.

\section{Related Work} \label{sec:related-work}
In this section, we first introduce some preliminaries of person ReID, and then present the related works about person ReID and federated lifelong learning. 

\subsection{Preliminary of Person ReID}
Person ReID can retrieve a person from non-overlapped camera views. 
The developments of deep neural network and the large-scale person ReID datasets have significantly improved the performances of person ReID in many vision-based applications \cite{leng2019survey, zhang2021global, shu2021spatiotemporal}.
Recently, many works \cite{leng2019survey, wang2018survey, shu2021large} focused on the data drift in real-world person ReID, and revealed that there exist spatial-temporal correlations for drifted data among different locations and moments.
The works that investigated the spatial-temporal correlations can be categorized into two lines.
One line of the works investigated the temporal correlations of data drift, which mainly focus on adopting domain adaptation \cite{li2021adadc, zhang2021cross} and lifelong learning \cite{pu2021lifelong, wang2016human, zhao2021continual, wu2021generalising} to narrow domain mismatches across different moments.
The other line of works  investigated the spatial correlations of data drift, which mainly adopt knowledge transfer \cite{wei2018person, zhuang2020performance} and feature alignment \cite{zhao2021incremental, zhao2022context, yang2022federated} to overcome the domain mismatches from different edges.

\subsection{Lifelong Person ReID}
In real-world person ReID deployments, camera environments and person characteristics are always different from that of training data. 
To narrow the domain shifts for the training and deployment stages, lifelong learning are preferred by some recent works \cite{pu2021lifelong, wang2016human, zhao2021continual, wu2021generalising, ge2022lifelong}.
Lifelong learning enables the models to continuously learn from new domains or scenarios, which has been widely adopted in many DNN-based serving systems \cite{huang2021modelci, zhang2020mlmodelci, zhang2021serverless, zhang2022towards} to deal with the domain drift.
The greatest challenge for lifelong person ReID is catastrophic forgetting, which requires ReID models to replay previous knowledge while continually training on new task streams. 
HVIL \cite{wang2016human} introduced a human-in-the-loop incremental learning method,  which enables models to adaptively refresh and optimize parameters by human feedback on unrecognized person images. 
AKA \cite{pu2021lifelong} proposed knowledge graph for lifelong person ReID, which can preserve the knowledge from previous domains while propagating learnt knowledge on unseen domains.
PTKP \cite{ge2022lifelong} cast lifelong person ReID as domain adaptation and proposed a pseudo task knowledge preservation framework to alleviate the domain gaps.
GwFReID \cite{wu2021generalising} proposed a class-imbalance lifelong learning for person ReID to generalise model representations for unseen domains.

\begin{figure*}[t]
  \centering
  \includegraphics[width=0.90\linewidth]{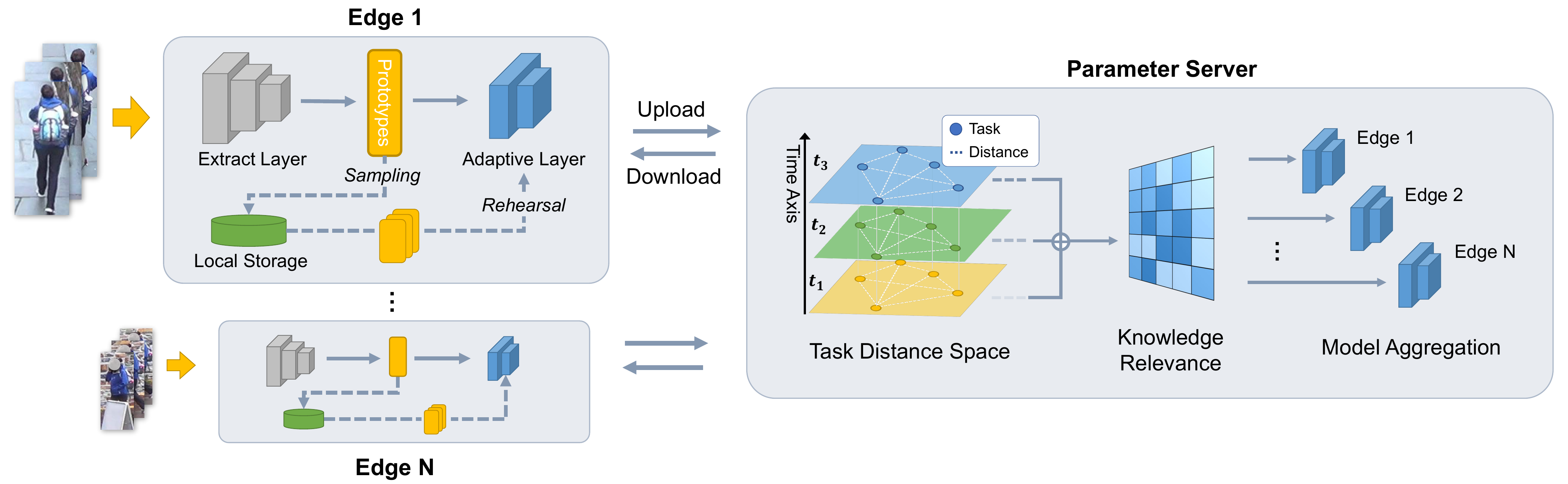}
  \caption{
  The architecture of FedSTIL for federated lifelong person ReID.
  The distributed edge clients continuously learn from both their local drift data and the relevant spatial-temporal knowledge from other edge clients organized by the parameter server to improve recognition accuracy.
  }
  \label{fig:architecture}
\end{figure*}

\subsection{Federated Person ReID}
With the growing data privacy concern, many person ReID systems are changing to the decentralized or federated training paradigm, 
where the private data are stored in isolated edges, and edges can jointly update models, instead of centralising private data for training. 
One challenge for federated person ReID is knowledge interference due to the data and domain heterogeneity across different edge clients.
To address this issue, FedDG \cite{yang2022federated} proposed domain and feature hallucinating techniques to train generalized person ReID models for federated learning with domain heterogeneity. 
FedPav \cite{zhuang2020performance} and some other works \cite{sun2021decentralised, zhang2021edge} adopted model aggregation and knowledge distillation to optimize the performances for the \textit{non-i.i.d} federated person ReID. 
FedReID \cite{wu2021decentralised} proposed an iterative client-server collaborative
learning to generalize person ReID domains without sharing private data.
SKA \cite{sun2021decentralised} proposed a selective knowledge aggregation method to transfer personalized knowledge among different edge clients. 
These methods are successful, but they did not consider the spatial-temporal correlation for domain knowledge from different edge locations and moments, which would limit the efficiency of knowledge sharing. 
Hence, our method aims to capture task-specific knowledge by integrating spatial-temporal knowledge to improve performances.

\subsection{Federated Lifelong Learning}
Despite the rapid progresses of lifelong person ReID and federated person ReID, few works studied lifelong person ReID under the federated learning paradigm, which we term as federated lifelong person ReID. 
Federated lifelong learning \cite{criado2022non, huang2022learn} can enable distributed models to collaboratively and continuously learn incremental knowledge from new scenarios, without sharing their private data.
To better explain the federated lifelong setting, we summarize the main research objectives in Table \ref{tab:setting_comparsion}, by comparing them with the lifelong setting and federal setting.
Specifically, federated lifelong person ReID has the merits of both lifelong and federated person ReID.
Besides, federated lifelong person ReID can also enable both spatial- and temporal- knowledge transfer \cite{shu2021large}, which is not simultaneously achievable for either the lifelong setting or the federal setting.
In this area, FedCurv \cite{shoham2019overcoming} and CFeD \cite{ijcai2022p303} mitigated the heterogeneity and forgetting by constraint training for \textit{non-i.i.d} federated lifelong learning.
FCIL \cite{dong2022federated} addressed the class-imbalanced and class-incremental federated lifelong learning by loss distillation.
FedWeIT \cite{yoon2021federated} proposed sparse parameters to reduce the network cost for federated lifelong learning.
These methods are successful, however, they are not specific designed for the person ReID scenario, and leaves a large room for improvement. 
For instance, FedCurv \cite{shoham2019overcoming} extended the regularization-based lifelong learning method \cite{kirkpatrick2017overcoming, aljundi2018memory} to the federated learning scenario, however, the high communication cost for transferring the regularization information is also an obstacle to apply it into the federated lifelong person ReID. 
FedWeIT \cite{yoon2021federated} utilized decomposed layers to reduce the network consumption.
It achieved great performance for many incremental learning scenarios \cite{van2018three}, where task IDs are provided.
However, it is not applicable in the scenarios where task IDs are unknown, such as the federated lifelong person ReID scenario. 
To the best of our knowledge, we are one of the first works which studied federated lifelong learning for person ReID.

\begin{table}[t]
    \caption{
    The comparison for three different person ReID settings.
    }
    \label{tab:setting_comparsion}
    \centering
    \begin{tabular}{l|c|c|c}
    \toprule
    \multirow{2.5}{3.2cm}{Research Objectives} & \multicolumn{3}{c}{Person ReID Settings}  \\
    \cmidrule{2-4}
     & Lifelong & Federal & Fed. + Life. \\
    \midrule
     Continual Training \cite{van2018three} & \checkmark & & \checkmark \\
     Data / Domain Shifts \cite{pu2021lifelong} & \checkmark & & \checkmark \\
     Model Forgetting Alleviation \cite{kirkpatrick2017overcoming} & \checkmark & & \checkmark \\
    \midrule
     Decentralised Training \cite{konevcny2016federated} & & \checkmark & \checkmark \\
     Data / Domain Heterogeneity \cite{yang2022federated} & & \checkmark & \checkmark \\
     Privacy Protection \cite{konevcny2016federated} & & \checkmark & \checkmark \\
    \midrule
     Temporal Knowledge Transfer \cite{shu2021large} & \checkmark & & \checkmark \\
     Spatial Knowledge Transfer \cite{shu2021large} & & \checkmark & \checkmark \\
    \bottomrule
    \end{tabular}
\end{table}

\begin{table}[h]
    \caption{Main Notations.}
    \label{tab:notation}
    \centering
    \begin{tabular}{ll}
    \toprule
    $\mathcal{D}_c^{(t)}$ & the drifted training data for edge client $c$ at round $t$  \\
    $\mathcal{P}_c^{(t)}$ & the compressed prototypes for raw data $\mathcal{D}_c^{(t)}$ \\
    $\overline{\mathcal{P}}_c^{(t)}$ & the task feature averaged on prototypes $\mathcal{P}_c^{(t)}$ \\ 
    \midrule
    $\mathcal{G}_c$ & the extraction layers for edge client $c$ \\
    $\mathcal{F}_c$ & the adaptive layers for edge client $c$ \\
    $\theta_c$ & the parameters of adaptive layers for edge client $c$ \\
    \midrule
    $A_c$ & the adaptive parameters with local knowledge \\
    $B_c$ & the base parameters with global spatial-temporal knowledge \\
    $\alpha_c$ & the attention to selectively capture the task-specific knowledge \\
    \midrule
    $\Pi(\cdot)$ & the similarity function to measure task features \\
    $\mathcal{S}_{ij}^{(t,t')}$ & the similarity between task feature $\overline{\mathcal{P}}_i^{(t)}$ and $\overline{\mathcal{P}}_j^{(t')}$ \\
    $W_{ij}^{(t)}$ & the knowledge relevance between $i$- and $j$-th edge client \\
    \midrule
    $A_{c}^{(r)}$ & the retrieval accuracy of edge client $c$ at the $r$-th round \\
    $F_{c}^{(r)}$ & the forgetting of edge client $c$ at the $r$-th round  \\
    \bottomrule
    \end{tabular}
\end{table}

\section{System Design} \label{sec:problem-definition}
In this section, we first present the problem definition of the federated lifelong person ReID, and then illustrate the system overview and learning procedure.
The main notations of the paper are illustrated in Table \ref{tab:notation}.

\subsection{Problem Definition}
We assume that the distributed person ReID system has $C$ edge clients. 
Each edge client $c$ continuously learns from its arriving task stream $\mathcal{D}_c^{(t)}$,
where $\mathcal{D}_c^{(t)}$ denotes the drift data arriving at the $t$-th round on the edge client $c$.
We assume that $\mathcal{D}_c^{(t)}$ is only available for edge client $c$ on the $t$-th round,
and the previous training data are no longer accessible due to the limited storage space of the edge clients. 
Our goal is to determine how to continuously learn from both on-edge task streams and across-edge knowledge to improve performances without sharing sensitive raw data among edge clients.

\begin{figure*}[t]
  \centering
  \includegraphics[width=0.90\linewidth]{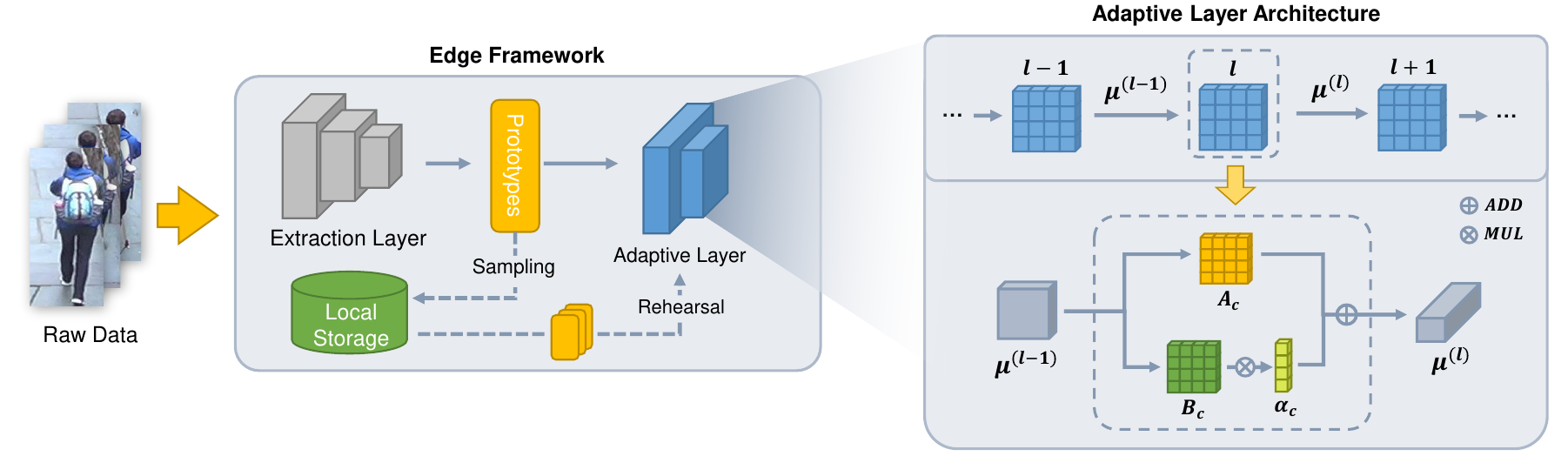}
  \caption{
    The architecture of the adaptive layers.
    The on-edge models continuously learn knowledge from the forthcoming tasks and meanwhile maintain the knowledge from prior tasks.
    The adaptive layers adaptively balance the tradeoff between local knowledge and global spatial-temporal knowledge.
  }
  \label{fig:adaptive_layer}
\end{figure*}

\subsection{System Overview}
As illustrated in Fig. \ref{fig:architecture}, our federated lifelong framework, FedSTIL, consists of one central parameter server and several distributed edge clients.
The edge clients continuously retrain local models with newly obtained drift data,
and the parameter server aggregates spatial-temporal knowledge from the edge clients' local models for knowledge sharing.
The network backbones are alternative, such as MobileNet, ResNet, and ViT. 
We divide the backbone of each edge model into two parts: 
1) the extraction layers initialized with global pre-trained weights to extract task prototypes, 
and 2) the adaptive layers for personalized lifelong learning. 
Considering the limited computing capacity of the edge clients \cite{wang2021smarteye, wang2022dynamic, dong2022collaborative}, our framework adopts the last several layers of the backbone as adaptive layers for training, and the rest layers are extraction layers.

The learning procedures of FedSTIL are as follows.
The edge clients collect the new drifted data as incremental tasks and extract the general representation from the drifted data as prototypes of raw data.
The edge clients learn incremental knowledge from these prototypes of drift data, and then upload learnt knowledge to the parameter server. 
The parameter server aggregates these incremental knowledge based on the spatial-temporal correlation of neighbors' task characteristics, and delivers the task-relevant knowledge to the edge clients. 
The edge clients adaptively utilize dispatched knowledge and previously learnt knowledge to optimize models for new scenarios continuously.
Meanwhile, some general prototypes of raw data will be stored in each edge client for future rehearsal to alleviate the forgetting of previously learnt knowledge.

\section{Learning Methodology} \label{sec:learning}
In this section, we first present the framework of our FedSTIL and then illustrate the training methodology.

\subsection{Lifelong Learning on Distributed Edges}
We enable the distributed edges to perform on-edge lifelong learning for every new scenario.
Considering the limited edge computing capacity, we only enable the last several layers, which we refer to as adaptive layers, for lifelong learning. 
The rest layers are pre-trained and frozen as the extraction layers.
In the real-world deployment, the drifted data from new scenarios are coming as a task stream, and the edge client will perform on-edge lifelong learning by the following procedures.

\begin{figure*}[t]
  \centering
  \includegraphics[width=0.95\linewidth]{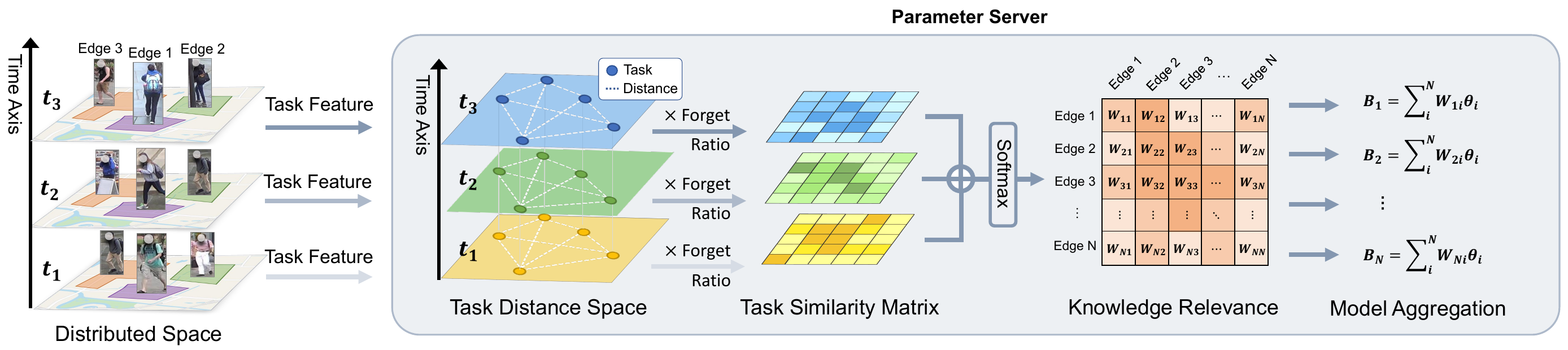}
  \caption{
  The framework of the spatial-temporal knowledge integration on the parameter server.
  The parameter server can automatically measure the spatial-temporal correlations for edge clients based on their task features.
  The task-specific knowledge is organized for edge clients for continuously learning.
  }
  \label{fig:integration}
\end{figure*}

\textbf{Task Prototype Extraction.}
The drift data arrives as task stream $\mathcal{D}_c^{(t)}$, where $t=0,1,2,...$.
Considering the limited resources of the edge clients, we utilize the extraction layers to encode the raw tasks $\mathcal{D}_c^{(t)}$ to compressed prototypes to represent the original task for training and inference.
Specifically, the drift data of edge client $c$ at the $t$-th round can be represented as $\mathcal{D}_c^{(t)}$ = $\{ (X_i^{(t)},Y_i^{(t)})\}$, where $X_i^{(t)}$ is the $i$-th training image of $\mathcal{D}_c^{(t)}$ and $Y_i^{(t)}$ is the corresponding label.
We input each raw data $(X_i^{(t)}, Y_i^{(t)}) \in \mathcal{D}_c^{(t)}$ into the extraction layers $\mathcal{G}_c$ to extract prototype as
\begin{equation}
\label{eqn:feature_extraction}
\begin{aligned}
\mathcal{P}_c^{(t)} = \{ ( \mathcal{G}_{c}(X_i^{(t)}), Y_i^{(t)} ) \},
\end{aligned}
\end{equation}
where $\mathcal{G}_{c}$ is the extraction layers, and $\mathcal{P}_c^{(t)}$ is the extracted prototype set for the raw data set $\mathcal{D}_c^{(t)}$.
Hence, the raw task $\mathcal{D}_c^{(t)}$ can be represented as the prototypes $\mathcal{P}_c^{(t)}$ after being processed by the extraction layers.
Compared with raw data, prototypes are more generalized to represent different tasks and also smaller to store.
Moreover, the prototypes with compressed semantics can also reduce data transmission costs and avoid potential privacy leakage.

\textbf{Adaptive Lifelong Learning.}
The training and inference tasks on different edge clients have different characteristics due to the different camera environment. 
If all of edge clients adopt one single unified model for lifelong training or inference, it is hard to achieve the optimal performance for each edge client. 

To address the problem of heterogeneous tasks on different edge clients, each edge client needs to have its personalized model, which can continuously learn from both local knowledge and other edge neighbors' knowledge.
Inspired by \cite{yoon2021federated}, we present the adaptive layers that leverage global and local knowledge for personalized model training. 
As illustrated in Fig. \ref{fig:adaptive_layer},
the parameters $\theta_c$ of the adaptive layers for the model of edge client $c$ consist of three parts: 
1) the adaptive parameters $A_c$ with knowledge learnt from local incremental tasks, 
2) the base parameters $B_c$ with the spatial-temporal knowledge from other correlated edge neighbors, 
and 3) the attention parameters $\alpha_c$ to capture the task-specific knowledge from the base parameters $B_c$ by attention mechanism,
\begin{align} \label{eqn:adaptive_layer}
\theta_c  = B_c \odot \alpha_c + A_c.
\end{align}

As shown in Eq. \eqref{eqn:adaptive_layer}, the parameters $\theta_c$ of the adaptive layers are combined by the following terms.
The first term ($B_c \odot \alpha_c $) enables learnable attention $\alpha_c$ for the spatial-temporal knowledge $B_c$ to capture the personalized specific knowledge for local tasks.
The second term ($A_c$) is to learn incremental knowledge from the drifted data of local tasks.
We adaptively combine knowledge from two different perspectives into the adaptive layers to continuously optimize models for the newly arriving tasks.

\begin{figure}[h]
  \centering
  \includegraphics[width=0.90\linewidth]{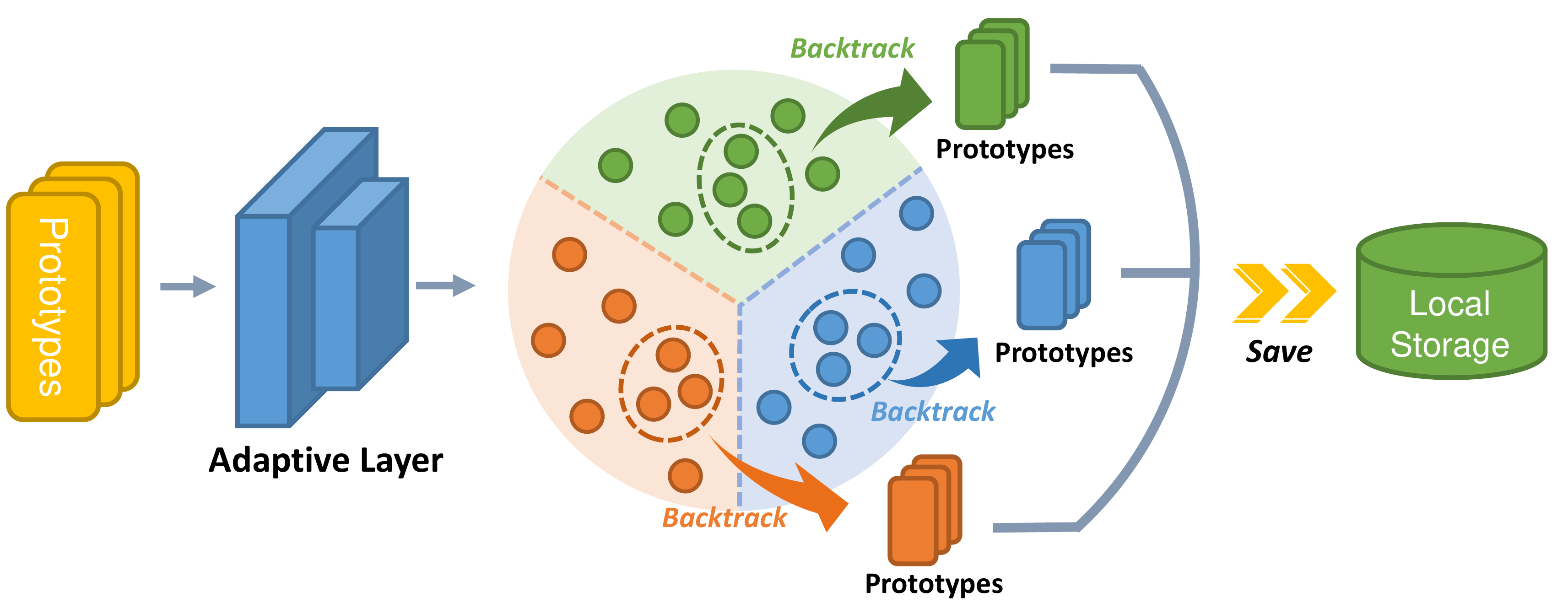}
  \caption{
  The data flow for sampling prototypes into local storage.
  We dynamically store some identities' prototypes for future rehearsal which are near the corresponding mean center.
  }
  \label{fig:rehearsal}
\end{figure}

\textbf{Prototype Rehearsal to Alleviate Forgetting.}
The local models of the edge clients are continuously updated with newly arriving tasks.
However, the new incremental knowledge will disturb the prior learnt knowledge, and thus the recognition accuracy for the previous domains will decrease during continuous learning \cite{wu2021generalising}. 
To alleviate catastrophic forgetting of previously learnt knowledge, we design a prototype rehearsal method which utilizes the stored general prototypes of the former tasks in the local storage for knowledge rehearsal.
As shown in Fig. \ref{fig:rehearsal}, 
we periodically sample some representative prototypes of the newly arriving incremental task with the nearest-mean-of-exemplars strategy \cite{rebuffi2017icarl}, and store them in the local storage of the edge clients. 
Specifically, when a training task arrives, we first input the prototypes of the task into the adaptive layers. 
Then, we calculate the mean center of the outputs of the adaptive layers for each person identity.
We will store some prototypes whose outputs are closest to the mean centers of different person identities.
Those prototypes are informative to maintain the representation of previously learnt knowledge.
During the training phases for new tasks, we periodically sample some stored prototypes for training to mitigate the forgetting of the previously learnt knowledge.

Compared with other rehearsal-based lifelong learning methods (e.g., iCaRL \cite{rebuffi2017icarl}, and GwFReID \cite{wu2021generalising}), our strategy is more friendly to edge devices with limited storage resources.
That is because prototypes are smaller than raw images for storage.
Meanwhile, rehearsing the prototypes to alleviate catastrophic forgetting takes smaller computation overhead.

\subsection{Spatial-Temporal Knowledge Integration on Server}
The distributed person ReID system consists of many edge clients.
However, not all knowledge learnt from edge clients are informative and relevant to others.
It may even hinder the training of an edge client if the irrelevant knowledge from others are transferred to the edge client \cite{yoon2021federated}. 
To extract the task-relevant knowledge from relevant neighbors, we design a spatial-temporal knowledge integration method based on the characteristics of the tasks from different locations and moments.
The framework of the spatial-temporal knowledge integration on the parameter server is illustrated in Fig. \ref{fig:integration}.
Next, we elaborate the procedure of spatial-temporal knowledge integration on the parameter server.

\textbf{Task Similarity across Spatial-Temporal Dimension.} 
The raw data of training task $\mathcal{D}_c^{(t)}$ is encoded into task prototypes $\mathcal{P}_c^{(t)}$,
which consist of the compressed semantics of the raw data.
To identify the spatial-temporal correlations of the edge tasks, however, directly centralizing and analyzing the raw data $\mathcal{D}_c^{(t)}$ or task prototypes $\mathcal{P}_c^{(t)}$ is consuming for communication, and exists the potential risk of privacy leakage.
Instead, the parameter server only collects the average value of the prototypes as the task feature $\overline{\mathcal{P}}_c^{(t)}$ for different edge tasks,
\begin{align} 
\overline{\mathcal{P}}_c^{(t)} = \frac{1}{|\mathcal{P}_c^{(t)}|} \sum_{p \in \mathcal{P}_c^{(t)}}{p},
\end{align}
where $\overline{\mathcal{P}}_c^{(t)}$ is the task feature for the training task on edge client $c$ at the $t$-th round,
$|\mathcal{P}_c^{(t)}|$ is the number of prototypes in the task, 
and $p$ is the vector of each prototype.

To evaluate the relevance of the tasks across different edge clients and moments,
we calculate the task similarity $\mathcal{S}_{ij}^{(t,t')}$ between the tasks of edge client $i$ at the $t$-th round and edge client $j$ at the $t'$-th round as
\begin{align} \label{eqn:tasksim}
\mathcal{S}_{ij}^{(t,t')} = \Pi(\overline{\mathcal{P}}_i^{(t)},\overline{\mathcal{P}}_j^{(t')}),
\end{align}
where $\Pi(\cdot)$ is the similarity function to measure between task feature $\overline{\mathcal{P}}_i^{(t)}$ and task feature $\overline{\mathcal{P}}_j^{(t')}$.
%
%
%
In this work, we adopt Kullback-Leibler Divergence as the similarity function because it can effectively measure the information difference between different distributions. 

\textbf{Knowledge Relevance for Distributed Edge Clients.} 
To organize the task-relevant knowledge for sharing, we calculate the knowledge relevance for distributed edge clients based on their task similarity across the spatial-temporal dimension.

We measure the knowledge relevance by the similarities of all historical tasks for different edge clients.
However, the tasks which arrive recently have a more significant impact on the model knowledge, 
while those coming earlier have less influence because the knowledge learnt from earlier tasks may be forgotten or updated over time.
Hence, we introduce the forgetting ratio $\lambda_f$ ($0 < \lambda_f < 1$) for historical task when calculating the knowledge relevance.
Formally, to calculate the knowledge relevance $W_{ij}^{(t)}$ between edge client $i$ and edge client $j$ at the $t$-th round,
we first evaluate the task similarity between the current task on edge client $i$ and the past $k$ tasks on edge client $j$.
Then, we accumulate the task similarities $\mathcal{S}_{ij}^{(t,t')}$ with the forgetting ratio as,
\begin{align}
\label{eqn:elevance}
W_{ij}^{(t)} = \sum_{t'=t-k}^{t} \lambda_{f}^{t-t'} \cdot \mathcal{S}_{ij}^{(t, t')}. 
\end{align}
Our intuition is that if the historical tasks of edge client $j$ have a higher similarity with the new task on edge client $i$, then the model knowledge of edge client $j$ has a greater relevance with the new task on edge client $i$.

\textbf{Personalized Model Aggregation.} 
The parameter server needs to integrate the task-relevant knowledge for each distributed edge client to improve the recognition accuracy.
%
%
%
We adopt the parameter-sharing approach for spatial-temporal knowledge transfer, 
and the parameters of models are the carrier of spatial-temporal knowledge for transferring.
The parameter server will aggregate the parameters of relevant edge models with corresponding knowledge relevance weight $W_{ij}^{(t)}$. 
Then the aggregated parameters will be dispatched to edge client $i$ as the base parameters $B_i$ for the training,
\begin{align}
\label{eqn:aggregate}
B_{i} = \sum_{j \in C/i} W_{ij}^{(t)} \cdot \theta_j,
\end{align}
where $\theta_j$ are the model parameters of edge client $j$, and $B_i$ are the task-specific base parameters for edge client $i$ with integrated knowledge from relevant edge clients.
The base parameters $B_i$ will be dispatched to edge client $i$ to help the optimization of the local model.

\begin{algorithm}[t]
    \DontPrintSemicolon
    \caption{Training Procedures of FedSTIL}
    \label{alg:algorithm1}
    \KwData{Pre-trained weights $ \theta_c^{(0)}$}
    \KwData{Task streams $\{ \mathcal{D}_c^{(t)} \}_{t=1}^{\infty}$}
    \KwData{Shared layers $\{\mathcal{G}_c\}_{c=1}^{C}$}
    \KwData{Adaptive layers $\{\mathcal{F}_c\}_{c=1}^C$}
    
    Initialize weights $\{ \alpha_c^{(0)}; A_c^{(0)} \}_{c=1}^C$\;
    \For{round $t=1,2,3,...$}{
        \For{client $c \in C$}{
            Collect incremental task $\mathcal{D}_c^{(t)}$ on edge client $c$\;
            
            \tcc{Task Prototype Extraction}
            Calculate task prototypes $\mathcal{P}_c^{(t)}$ and $\overline{\mathcal{P}}_c^{(t)}$\;
            
            \tcc{Spatial-Temporal Integration}
            Calculate task similarity $\{\mathcal{S}_{ci}^{(t,t')}\}_{i\in C/c}$ by Eq.\eqref{eqn:tasksim}\;
            Integrate knowledge relevance $\{W_{ci}^{(t)}\}_{i\in C/c}$ by Eq.\eqref{eqn:elevance}\;
            Aggregate spatial-temporal knowledge into $B_c$ by Eq.\eqref{eqn:aggregate}\;
            
            \tcc{Adaptive Lifelong Learning}
            Set parameters of local adaptive layers $\theta_c \leftarrow B_c \odot \alpha_c + A_c $\;

            \tcc{Prototypes Rehearsal}
            Sample training data $X$ from current and stored prototypes\;
            \For{ epoch = $1,2,...$ }{
                Update $\theta_c \leftarrow \theta_c - \eta \nabla \mathcal{L} (\theta_c; X) $
            }
        }
        Upload parameters $\theta_c$ to the parameter server\;
    }
\end{algorithm}

\subsection{Training Methodology}
The distributed training procedures of FedSTIL are illustrated in Algorithm \ref{alg:algorithm1}. 
The edge clients collect new drifted data as new tasks, and input these tasks into their extraction layers to generate feature vectors as the prototypes.
Then, the edge clients upload the task features, which are the average of the prototypes, to the parameter server.
The parameter server calculates the similarities with the task features of the historical tasks from different edge clients.
Based on these similarities, the parameter server will integrate the task-relevant knowledge as the base parameters for each edge client, and the edge clients can continuously learn from the new tasks based on the integrated knowledge from other edge clients. 
Finally, when the training of the edge clients' models converges, the parameters of the adaptive layers of the edge models will be uploaded to the parameter server to further improve other spatial-temporal correlated edge clients.

For each task, the edge clients store some of the task prototypes for future rehearsal, which can alleviate forgetting of previously learnt knowledge. 
Specifically, we periodically select a batch of prototypes from the current task and previously stored prototypes to train the parameters $\theta_c$ of the adaptive layers for each task. 
The loss function can be either cross-entropy loss or triplet loss.

The training samples on the edge clients may be insufficient due to the limited data.
If we directly use the limited data to train models, the models could easily fall into overfitting. 
Hence, we adopt parameter tying to tackle this issue by regularizing or penalizing model weights \cite{zhang2018learning}, 
where all parameter changes are summarized as a penalty loss to get sparse gradients for parameters optimization.
By tying the parameters of edge clients' models, the models can converge with less overfitting due to the minimal changes in prior knowledge.

\section{Experiment} \label{sec:experiment}
In this section, we present the experimental settings and evaluate the performances of our method by comparing with the baseline methods.
We also conduct ablation and visualization studies to validate the effectiveness of our method.

\subsection{Experimental Setting}

\subsubsection{Dataset} \label{subsec:dataset}
\begin{figure}[t]
  \centering
  \includegraphics[width=\linewidth]{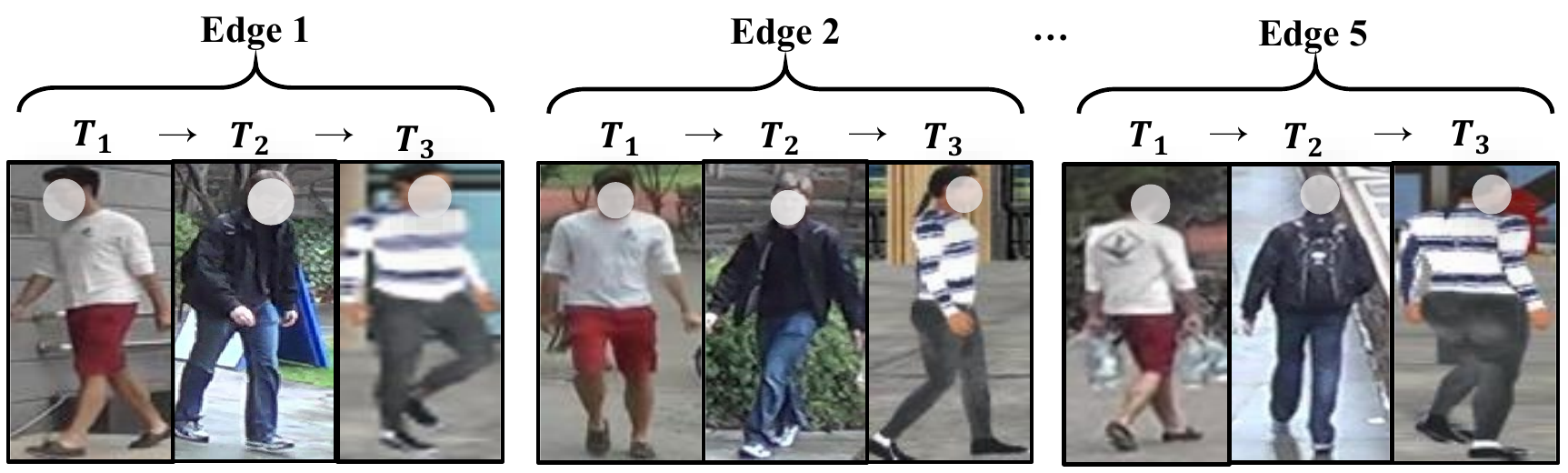}
  \caption{
  The visualization of some examples in our datasets. We maintain the distribution gaps among edge clients with different camera views.
  }
  \label{fig:dataset_vis}
\end{figure}

\begin{figure}[t]
\centering
\subfigure[Hard-divided Datasets (Prior)]{ 
\includegraphics[width=0.48\linewidth]{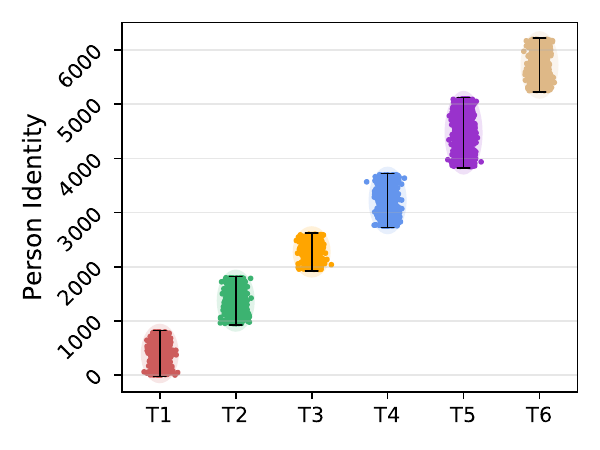}\label{fig:datasets_a}}
\subfigure[Soft-divided Datasets (Ours)]{
\includegraphics[width=0.48\linewidth]{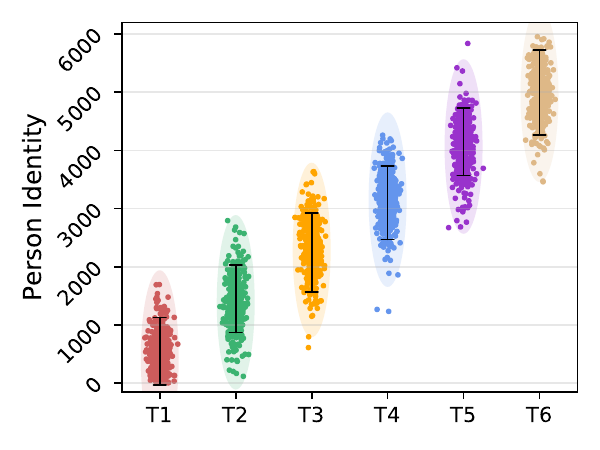}\label{fig:datasets_b}}
\caption{
Person identities distribution at different moments. We adopt soft-divided datasets to enable the data distribution changes gradually. 
}
\label{fig:datasets}
\end{figure}

To simulate the real-world scenarios of spatial-temporal person ReID \cite{shu2021large}, we apply the benchmark combined from five popular person ReID datasets \footnote{\href{https://github.com/MSNLAB/ST-ReID-Datasets}{https://github.com/MSNLAB/ST-ReID-Datasets}}.
In this benchmark, we split the images into five partitions, and each partition has the person images of different camera-IDs.
Then, we shuffle these partitions into five distributed edge clients.
As shown in Fig. \ref{fig:dataset_vis}, we ensure the person images of different edge clients are from different camera views, which aims to align with the real-world camera distributions \cite{zheng2015scalable}.
Then, we group the images of each edge client into 6 sequential tasks, and each edge sequential task uses 60\% of the images as training data and 40\% as query data.
The gallery images are organized from other edges' query images, where we can ensure the camera-IDs of gallery images are different from query images.
Additionally, we simulate the data drift on an edge client by changing person identities distribution.
As shown in Fig.~\ref{fig:datasets_a}, the data drift (e.g., T1 → T2) in prior efforts can be measured as the class shift (e.g., Person-ID: 0-800 → 800-1600).
However, in the real-world scenario, the person identities between adjacent moments always change gradually rather than completely.
Therefore, as shown in Fig.~\ref{fig:datasets_b}, we enable the person identities to change gradually in the form of normal distribution, where some old and new identities can be all involved during one moment \cite{van2018three}

\subsubsection{Backbone Network}
The backbone network for person ReID is modified from ResNet-18 \cite{luo2019bag}. 
Compared with regular ResNet-18, the differences are as follows:
1) the last stride of backbone is set to be 1 to enrich representation granularity;
%
2) a batch normalization layer is added after representation for better convergence;
%
3) the bias of the classifier is removed to avoid representation bias.
%
To verify the compatibility of our method with different backbones, we also evaluate the performances with ResNet-50 and Swin-Transformer \cite{liu2021swin}.

\subsubsection{Training Setting}
We adopt Adam optimizer with learning rate $10^{-3}$ and weight decaying rate $10^{-5}$. 
The edge models are trained for 5 epochs at each communication round between parameter server and edge clients. 
To avoid the overfitting issue, the training phases will stop if the loss stops decreasing for 3 epochs.
%
%
Considering the limited computing capacity of the edge devices, only the last residual block and the classifier will be updated during training, and the other layers are fixed with pre-trained weights.

\subsubsection{Performance Metric}
We adopt the following performance metrics to evaluate federated lifelong person ReID, which are in line with the previous works \cite{yoon2021federated, leng2019survey}. 

\begin{itemize}
\item[(1)] 
Accuracy: we adopt the mean average precision (mAP) and cumulative match characteristic (CMC) \cite{leng2019survey} to measure the retrieval accuracy for each edge client. 
The test accuracy (i.e., mAP, CMC) at communication round $r$ is defined as the average retrieval accuracy of all training tasks on edge client $c$ as follows,
\begin{small}
\begin{align}\small\label{equ:macc}
A_c^{(r)} = \frac{1}{N_c}\sum_{i=1}^{N_c}a(r;\mathcal{D}_c^{(i)}),
\end{align}
\end{small}

where $a(r;\mathcal{D}_c^{(i)})$ is the retrieval accuracy of the $i$-th task $\mathcal{D}_c^{(i)}$ on edge client $c$ at $r$-th communication round and $N_c$ is the number of training tasks for edge client $c$.

\item[(2)] 
Forgetting: we measure the forgetting for each client by calculating the decreasing accuracy compared with the maximum value of each task during training \cite{yoon2021federated}.
The forgetting of client $c$ can be calculated as follow,
\begin{small}
\begin{align}\label{eqn:mfor}
F_c^{(r)} = \frac{1}{N_c-1} \sum_{i=1}^{N_c-1}
           \max_{t \in \{1,...,r\}} a(t;\mathcal{D}_c^{(i)})-a(r;\mathcal{D}_c^{(i)}),
\end{align}
\end{small}

where $a(r;\mathcal{D}_c^{(i)})$ is the retrieval accuracy as mentioned above.
Note that the forgetting for the last training task does not exist because there is no further task for training.
\end{itemize}

\begin{table*}[t]
    \caption{
    The performance comparison of different methods. 
    }
    \label{tab:results}
    \centering
    \setlength{\tabcolsep}{3.6mm}{
    \begin{tabular}{c|c|cccc|c|cc} 
    \toprule
    \textbf{Type} & \textbf{Method} &  \textbf{mAP \small{(\%)}} & \textbf{R1 \small{(\%)}} & \textbf{R3 \small{(\%)}} & \textbf{R5 \small{(\%)}} & \textbf{Storage} & \textbf{S2C} & \textbf{C2S} \\
    \midrule
    \multirow{1}{1.8cm}{\centering Baseline} &
    STL & 54.39 & 51.62 & 60.27 & 63.70 & 354MB & NaN & NaN \\
    \midrule
    \multirow{2}{1.8cm}{\centering Lifelong (Regularization)} & 
    EWC \cite{kirkpatrick2017overcoming} & 52.42 & 49.56 & 58.34 & 61.87 & 154MB & NaN & NaN \\
    & MAS \cite{aljundi2018memory} & 52.32 & 49.35 & 58.53 & 62.31 & 154MB & NaN & NaN \\
    \midrule
    \multirow{5}{1.8cm}{\centering Lifelong (Rehearsal)} & 
    LwF \cite{li2017learning} & 52.19 & 49.13 & 58.02 & 61.65 & 106MB & NaN & NaN \\
    & GwF \cite{wu2021generalising} & 52.67 & 49.87 & 58.49 & 61.87 & 287MB & NaN & NaN \\
    & iCaRL \cite{rebuffi2017icarl} & 54.52 & 51.78 & 60.33 & 63.63 & 696MB & NaN & NaN \\
    & AKA \cite{pu2021lifelong} & 55.02 & 52.23 & 60.75 & 64.10 & 178MB & NaN & NaN \\
    & PTKP \cite{ge2022lifelong} & 55.22 & 52.53 & 60.90 & 64.15 & 395MB & NaN & NaN \\
    \midrule
    \multirow{3}{1.8cm}{\centering Federated} & 
    FedSGD \cite{konevcny2016federated} & 60.53 & 57.68 & 66.42 & 69.07 & 59MB & 2.8GB & 2.8GB \\
    & FedAvg \cite{konevcny2016federated} & 62.47 & 60.26 & 67.20 & 69.92 & 59MB & 2.8GB & 2.8GB \\
    & FedProx \cite{li2020federated} & 62.48 & 60.26 & 67.30 & 69.97 & 106MB & 2.8GB & 2.8GB \\
    \midrule
    \multirow{9}{1.8cm}{\centering Federated Lifelong} & 
    FedCurv \cite{shoham2019overcoming} & 58.26 & 55.42 & 64.37 & 67.75 & 631MB & 30.7GB & 6.1GB \\
    & CFeD \cite{ijcai2022p303} & 60.09 & 57.71 & 65.28 & 68.02 & 378MB & 2.8GB & 2.8GB \\
    & FedWeIT (a) \cite{yoon2021federated} & 53.86 & 50.93 & 59.92 & 63.67 & 611MB & 3.2GB & 1.1GB \\
    & FedWeIT (b) \cite{yoon2021federated} & 64.06 & 61.55 & 68.89 & 71.38 & 993MB & 8.1GB & 2.7GB \\
    \cmidrule{2-9}
    & EWC \cite{kirkpatrick2017overcoming} + FedAvg \cite{konevcny2016federated} & 53.16 & 50.29 & 59.14 & 62.66 & 154MB & 2.8GB & 2.8GB\\
    & EWC \cite{kirkpatrick2017overcoming} + FedProx \cite{li2020federated} & 53.60 & 50.73 & 59.69 & 63.18 & 154MB & 2.8GB & 2.8GB \\
    & iCaRL \cite{rebuffi2017icarl} + FedAvg \cite{konevcny2016federated} & 59.98 & 57.59 & 65.14 & 67.84 & 758MB & 2.8GB & 2.8GB \\
    & iCaRL \cite{rebuffi2017icarl} + FedProx \cite{li2020federated} & 60.55 & 58.32 & 65.35 & 67.94 & 612MB & 2.8GB & 2.8GB \\
    \cmidrule{2-9}
    & \textbf{FedSTIL (ours)} & \textbf{68.16} & \textbf{66.05} & \textbf{72.03} & \textbf{74.05} & \textbf{825MB} & \textbf{2.8GB} & \textbf{2.8GB} \\
    \bottomrule
    \end{tabular}}
\end{table*}

\begin{figure*}[t]
  \centering
  \subfigbottomskip=2pt
  \subfigcapskip=-8pt
  \subfigure[mAP Accuracy]{
    \includegraphics[width=\linewidth]{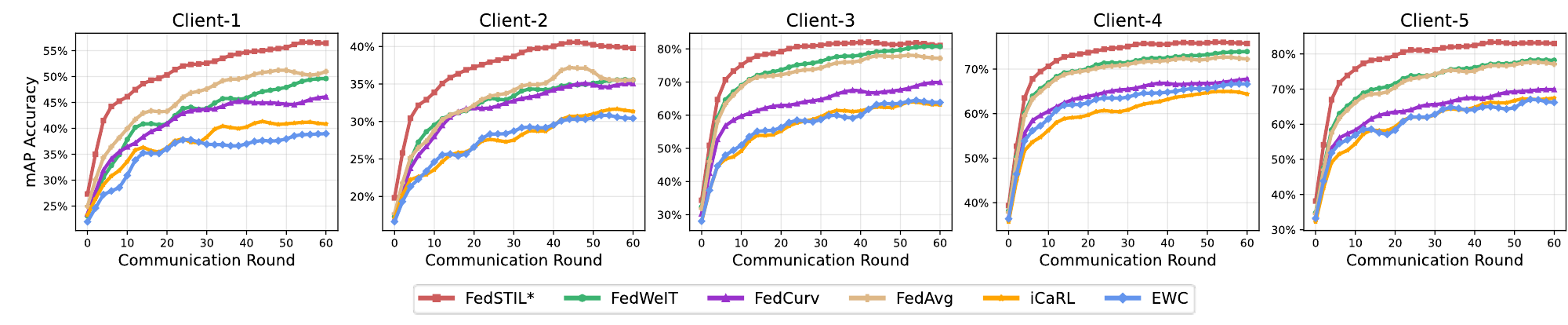}} \\
  \subfigure[Rank-1 Accuracy]{
    \includegraphics[width=\linewidth]{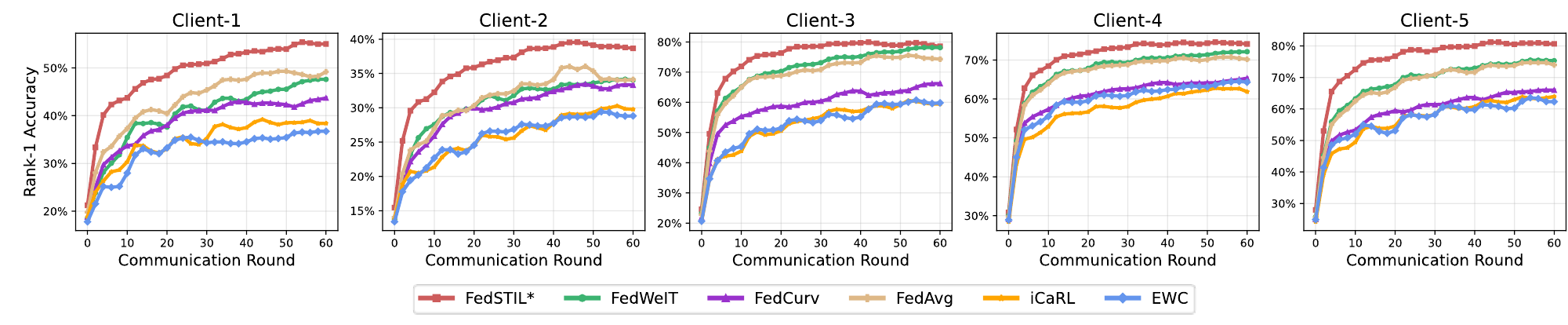}} \\
  \caption{
  The comparison of average mAP and Rank-1 accuracy of 6 local tasks on 5 edge clients during 60 communication rounds.
  Our FedSTIL achieves higher and stable accuracy with less fluctuation during the federated lifelong learning.
  }
  \label{fig:accuracy_detail}
\end{figure*}

\subsection{Performance Comparison}
We compare the accuracy (mAP, Rank-1, Rank-3, Rank-5), storage cost (model size + memory size), server-to-client (S2C) and client-to-server (C2S) communication cost of our method with the following baseline methods:
1) Single task learning (STL) with multi-models for all tasks; 
2) Lifelong learning: EWC  \cite{kirkpatrick2017overcoming}, MAS \cite{aljundi2018memory}, iCaRL \cite{rebuffi2017icarl}, LwF \cite{li2017learning}, GwF \cite{wu2021generalising}, AKA \cite{pu2021lifelong}, and PTKP \cite{ge2022lifelong}; 
3) Federated learning: FedSGD \cite{konevcny2016federated}, FedAvg \cite{konevcny2016federated}, and FedProx \cite{li2020federated}; 
4) Federated lifelong learning: CFeD \cite{ijcai2022p303},  FedCurv \cite{shoham2019overcoming}, FedWeIT \cite{yoon2021federated}, and some combinations of federated and lifelong methods.

In the real-life deployments of person ReID, task IDs are unavailable both in the training and inference stages. 
FedWeIT \cite{yoon2021federated} requires task IDs for training and inference, and we assume that task IDs are given for FedWeIT \cite{yoon2021federated} to obtain its performance.
Moreover, we validate FedWeIT \cite{yoon2021federated} under different settings to balance its accuracy and communication cost. FedWeIT (a) is set with $l_1=1.0\times10^{-4}$, $l_2=1.0\times10^{-6}$. FedWeIT (b) is set with $l_1=5.0\times10^{-6}$, $l_2=1.0\times10^{-3}$.

Table \ref{tab:results} illustrates the performance comparison of different methods. 
Our FedSTIL can outperform other baseline methods in accuracy and communication cost.
The most competitive baseline method, FedWeIT \cite{yoon2021federated}, is about 4\% lower than our method on mAP, and the S2C communication cost is 2.89 times higher compared with our method.
The higher accuracy and lower communication cost make our method more applicable in real-world person ReID scenarios. 
The detailed analysis for each performance metric are as follows.

\subsubsection{Comparison of Accuracy}
We evaluate the average accuracy of different edge clients under different federated lifelong learning methods (i.e., FedSTIL, FedWeIT \cite{yoon2021federated}, FedCurv \cite{shoham2019overcoming}, and CFeD \cite{ijcai2022p303}) in Fig \ref{fig:accuracy_detail}.
Compared with the baseline methods, our method can achieve higher accuracy over 60 rounds of training.
Besides, with continuous training on the new tasks, the accuracy can increase stably with less fluctuation, which also indicates the robustness of our methods.
This improvement is largely owing to the effective knowledge sharing and the alleviation of the forgetting of the learnt knowledge during federated lifelong learning. 
We then analyze the impact of knowledge sharing and forgetting on federated lifelong person ReID.

\begin{figure*}[t]
  \centering
  \includegraphics[width=\linewidth]{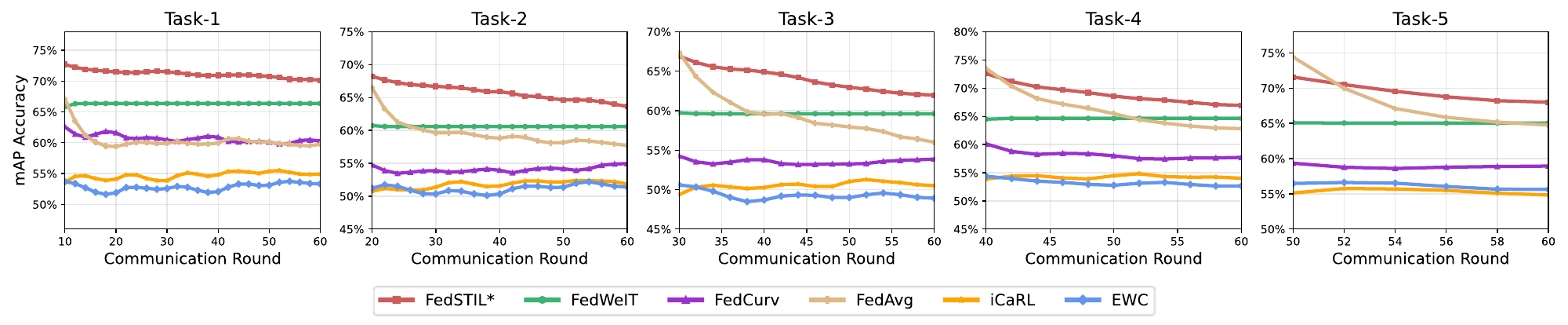}
  \caption{
  The comparison of forgetting for trained tasks over different communication rounds.
  Our FedSTIL can maintain higher accuracy for ever-learnt tasks with less forgetting when training for new tasks.
  }
  \label{fig:forgetting}
\end{figure*}

\textbf{Knowledge Sharing.} 
Our FedSTIL can achieve higher accuracy because it can effectively exchange knowledge among edge clients.
In general, most federated learning-based methods (i.e., FedSGD \cite{konevcny2016federated}, FedAvg \cite{konevcny2016federated}, FedProx \cite{li2020federated},
CFeD \cite{ijcai2022p303}, FedCurv \cite{shoham2019overcoming}, FedWeIT \cite{yoon2021federated}, and our FedSTIL) can achieve higher accuracy than other local-based training methods (i.e., STL, EWC \cite{kirkpatrick2017overcoming}, MAS \cite{aljundi2018memory}, iCaRL \cite{rebuffi2017icarl}, LwF \cite{li2017learning}, GwF \cite{wu2021generalising}, AKA \cite{pu2021lifelong}, and PTKP \cite{ge2022lifelong}). 
These results indicate that exchanging on-edge knowledge can significantly improve the overall accuracy,
because the local data of an edge client is limited for learning, such as insufficient data or limited camera views.
Thus, sharing the knowledge across edge clients can make up for the limitations of local data.
Moreover, we can observe that our FedSTIL achieves the highest accuracy compared with other federated learning-based methods (i.e., FedSGD \cite{konevcny2016federated}, FedAvg \cite{konevcny2016federated}, FedProx \cite{li2020federated}, CFeD \cite{ijcai2022p303}, FedCurv \cite{shoham2019overcoming}, and FedWeIT \cite{yoon2021federated}).
The improvement mainly owes to the task-relevant knowledge sharing among edge clients, which alleviates the interference incurred by the irrelevant knowledge transferring among the edge clients' models.

\begin{figure}[h]
  \centering
  \includegraphics[width=\linewidth]{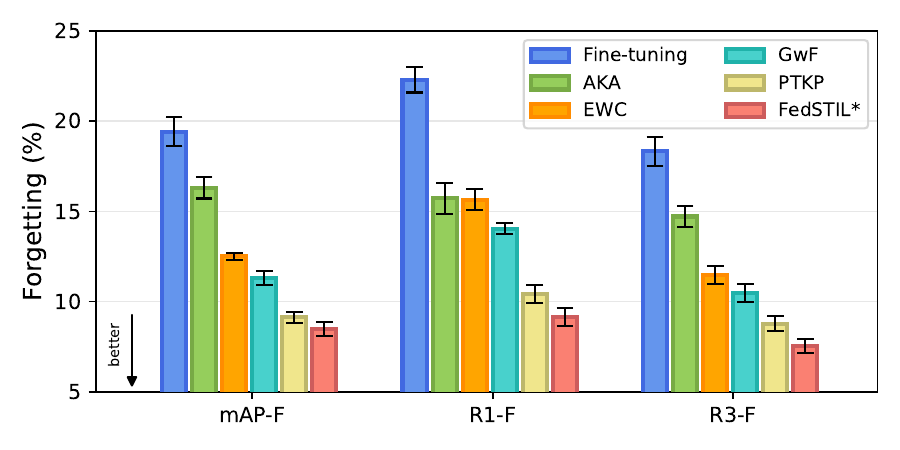}
  \caption{
  The comparison of the catastrophic forgetting  under the MSMT17 dataset. Our FedSTIL alleviates the forgetting under lifelong learning for new scenarios.
  }
  \label{fig:forgeting_msmt}
\end{figure}

\textbf{Alleviation of Forgetting.} 
Our FedSTIL can maintain the accuracy with less forgetting of the previous knowledge during the continuous training on the subsequent tasks. 
We can observe in Fig. \ref{fig:accuracy_detail} that both the Rank-1 and mAP accuracy of federated learning methods (i.e., FedAvg \cite{konevcny2016federated} and FedProx \cite{li2020federated}) will not increase after the 40$^{th}$ round. 
As shown in Fig. \ref{fig:forgetting}, this is because the knowledge learnt from the past tasks is gradually forgotten during the learning of the new tasks. 
In contrast, federated lifelong methods (i.e., FedCurv \cite{shoham2019overcoming}, FedWeIT \cite{yoon2021federated}, and FedSTIL) can alleviate forgetting on the past tasks, which help to achieve higher accuracy.
To future investigate the effectiveness of our methods, we compare catastrophic forgetting with the fine-tuning baseline (without lifelong learning) and some other lifelong learning methods on MSMT17 dataset \cite{wei2018person}.
Also, we divide the MSMT17 dataset into 6 task sequences, and the model backbone is ResNet50.
Fig.~\ref{fig:forgeting_msmt} shows us the forgetting after finishing lifelong learning on the MSMT17 dataset.
We can observe that our method FedSTIL can still effectively alleviate the forgetting compared with other lifelong person ReID baselines.

\subsubsection{Comparison of Communication Cost}

\begin{figure}[t]
  \centering
  \includegraphics[width=0.90\linewidth]{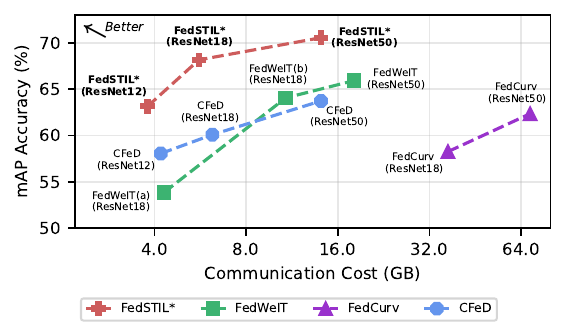}
  \caption{
  The comparison of mAP accuracy over communication cost.
  Our FedSTIL achieves the highest retrieval accuracy with less communication cost.
  }
  \label{fig:communication}
\end{figure}

We illustrate the mAP accuracy over total communication costs (client-to-server and server-to-client cost) in Fig. \ref{fig:communication}.
FedSTIL is communication-efficient and can achieve higher accuracy compared with the other federated lifelong learning methods.
FedWeIT \cite{yoon2021federated} can adjust the communication cost by $l_1$ pruning for the transferred parameters, however, the accuracy of FedWeIT (a) is still much lower than our method under comparable communication cost.
FedCurv \cite{shoham2019overcoming} has lower accuracy with higher communication cost.
That is because FedCurv \cite{shoham2019overcoming} needs to transfer excessive additional information to alleviate catastrophic forgetting. 
Our FedSTIL is more communication-efficient because it does not need to exchange additional parameters among edge clients except model weights. 
This improvement is largely because FedSTIL only requests task-specific parameters, which have been aggregated by the parameter server based on the relevant spatial-temporal knowledge.  
Moreover, our FedSTIL can achieve higher accuracy compared with the other baseline methods by using the cheaper model ResNet18, which has lower computation cost and storage consumption.
Therefore, FedSTIL is more applicable for deployment in real-life person ReID scenarios by reducing communication and computational costs.

\begin{table}[h]
  \caption{
  The ablation study of the influences of edge and server modules on the accuracy of our method.
  }
  \label{tab:eff_accuracy}
  \centering
  \begin{tabular}{lcc}
    \toprule
    Variant & mAP \small{(\%)} & R1 \small{(\%)} \\
    \midrule
    \textbf{FedSTIL} & \textbf{68.16} & \textbf{66.05} \\
    \midrule
    \textbf{Edge Modules} \\
    \quad \textit{w/o} Adaptive Layer & 63.71 (-4.45) & 62.54 (-3.51) \\
    \quad \textit{w/o} Prototype Rehearsal & 60.73 (-7.43) & 58.18 (-7.87) \\
    \quad \textit{w/o} Parameter Tying & 62.53 (-5.63) & 60.33 (-5.72) \\
    \midrule
    \textbf{Server Modules} \\
    \quad \textit{w/o} Spatial Model Aggregation & 58.72 (-9.44) & 57.03 (-9.02) \\
    \quad \textit{w/o} Temporal Model Aggregation & 61.04 (-7.12) & 59.53 (-6.52) \\
    \quad \textit{w/o} Spatial-Temporal Integration & 54.26 (-13.90) & 51.51 (-14.54) \\
  \bottomrule
\end{tabular}
\end{table}

\subsection{Ablation Study}
We conduct the ablation study by removing some key components of our method to analyze their influences on the performances, and the results are shown in Table \ref{tab:eff_accuracy}.
Specifically, we first remove the edge modules to verify our lifelong learning effectiveness.
The "\textit{w/o} Adaptive Layer" setting replaces our adaptive layer with standard convolution kernels and MLP blocks.
The "\textit{w/o} Prototypes Rehearsal" setting eliminates the rehearsal of historical prototypes in lifelong learning.
The "\textit{w/o} Parameter Tying" setting removes the tying regularization terms in the loss functions.
Moreover, we also verify the distributed knowledge sharing by removing the parameter server modules.
Specifically, the "\textit{w/o} Spatial Model Aggregation" setting removes the model aggregation on the spatial-relevant edges.
The "\textit{w/o} Temporal Model Aggregation" setting omits the historical task features when calculating task distances in Eq. \eqref{eqn:elevance}.
The "\textit{w/o} S-T Integration" setting removes both spatial- and temporal- knowledge sharing among edge clients.
Table \ref{tab:eff_accuracy} demonstrates that mAP and Rank-1 accuracy will decrease significantly without these modules, which also verifies that these modules can contribute to our method's performance improvements. 
We then conduct detailed ablation studies to verify the effectiveness of each design of our FedSTIL.

\begin{table}[h]
  \caption{The comparison of catastrophic forgetting with different memory sizes for prototypes rehearsal.}
  \label{tab:forgetting}
  \centering
  \setlength{\tabcolsep}{2mm}{
  \begin{tabular}{lcccc}
    \toprule
    Variant & Memory & mAP-F ($\downarrow$) & R1-F ($\downarrow$)  & R5-F ($\downarrow$) \\
    \midrule
    \textit{w/o} PR & NaN & 5.40 & 5.71 & 4.26 \\
    \midrule
    \quad+ PR: 10K & 415MB & 4.87 & 5.25 & 3.51 \\
    \quad+ PR: 12K & 505MB & 4.20 & 4.59 & 3.19 \\
    \quad+ PR: 14K & 572MB & 4.02 & 4.36 & 2.76 \\
    \quad+ PR: 16K & 662MB & 4.14 & 4.48 & 2.92 \\
    \quad+ PR: 18K & 719MB & 3.78 & 4.03 & 2.83 \\
    \quad+ PR: 20K & 783MB & 3.57 & 3.81 & 2.75 \\
  \bottomrule
\end{tabular}}
\end{table}

\textbf{Influence of Memory Size on Forgetting.} 
We analyze the catastrophic forgetting of our method as Eq. \eqref{eqn:mfor} under different memory sizes for prototype rehearsal.
As shown in Table \ref{tab:forgetting}, the Rank-1 Forgetting (R1-F), Rank-5 Forgetting (R5-F), and mAP Forgetting (mAP-F) keep decreasing as the memory size increases.
With a larger memory size for storing prototypes, the knowledge from more historical tasks can be saved for future rehearsal.
Therefore, the rehearsal with more historical prototypes can effectively alleviate the forgetting of the previous knowledge. 
In addition, the Rank-1 Forgetting with the memory size 20,000 (20K) is nearly $2\%$ lower than without prototype rehearsal, which is within the acceptable range in the real-life person ReID scenario. 
Additionally, we also observe that forgetting decreases slowly or even rebounds (e.g., PR: 16K) as we give more memory size.
The most probable reason is that some stored prototypes may be redundant or even noisy with larger memory space.
In other words, although we store more prototypes, we can hardly gain equivalent knowledge for future rehearsal.
Hence, one needs to carefully pick up the PR threshold to balance the tradeoff between memory usage and knowledge forgetting.
Nevertheless, these results can still indicate that our prototype rehearsal effectively alleviate the catastrophic forgetting and retain the knowledge from the previous tasks during federated lifelong learning.

\begin{table}[h]
  \caption{
  Performances with different backbones.
  }
  \label{tab:backbone}
  \centering
  \begin{tabular}{lcccccc}
    \toprule
    Backbone & Method & mAP & R1 & R5 & Storage & TC \\
    \midrule
    \multirow{5}{*}{ResNet18} 
    & 
    FedCurv & 58.26 & 55.42 & 67.75 & 631MB & 36.8GB \\
    & CFeD & 60.09 & 57.71 & 68.02 & 378MB & 5.6GB \\
    & FedWeIT & 64.06 & 61.55 & 71.38 & 993MB & 10.8GB \\
    & FedSTIL & 68.16 & 66.05 & 74.05 & 825MB & 5.6GB \\
    \midrule
    \multirow{5}{*}{ResNet50} 
    & 
    FedCurv & 62.35 & 58.72 & 70.03 & 951MB & 68.5GB \\
    & CFeD & 63.73 & 59.31 & 72.14 & 508MB & 14.1GB \\
    & FedWeIT & 65.92 & 62.04 & 73.58 & 1.42GB & 18.1GB \\
    & FedSTIL & 70.55 & 68.30 & 75.14 & 1.62GB & 14.1GB \\
    \midrule
    \multirow{5}{*}{Swin-T}
    & 
    FedCurv & 65.71 & 62.30 & 72.95 & 1.14GB & 92.2GB \\
    & CFeD & 65.58 & 62.42 & 72.08 & 703MB & 15.8GB \\
    & FedWeIT & 66.52 & 64.35 & 74.70 & 1.90GB & 22.5GB \\
    & FedSTIL & 71.31 & 69.12 & 75.03 & 1.92GB & 15.8GB \\
  \bottomrule
\end{tabular}
\end{table}

\textbf{Performances with Different Backbones.} 
We analyze the influences of choosing different network backbones on the accuracy, storage cost, and communication cost.
Specifically, we utilize ResNet18, ResNet50, and Swin-Transformer (Swin-T) \cite{liu2021swin} as backbones for training.
Considering the limited computing capacity of edge devices, only the last block of these backbones are trainable during learning, and the rest blocks are fixed with the pre-trained weights.
As shown in Table \ref{tab:backbone}, our FedSTIL can outperform the other baselines on accuracy (i.e., mAP, Rank-1, Rank-5), storage, and total communication cost (TC) with different backbones.
The storage size of FedSTIL is comparable with FedWeIT \cite{yoon2021federated} and larger than FedCurv \cite{shoham2019overcoming}.
The communication cost of FedSTIL is lower than other baselines, while the accuracy (e.g., mAP, Rank-1, and Rank-5) of FedSTIL is the highest with different backbones. 
These results verify the compatibility of our method with different backbones.
Therefore, one can choose suitable backbones based on the edge computing capacity.

\begin{figure*}[t]
  \centering
  \includegraphics[width=\linewidth]{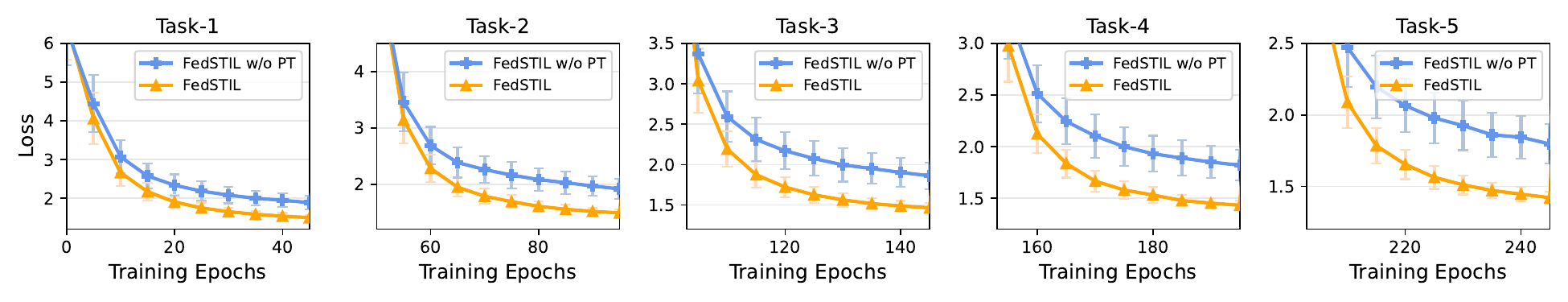}
  \caption{
  The comparison of the averaged loss for different tasks w/ and w/o parameter tying.
  Parameter Tying can achieve convergence faster under continuously training for sequential tasks.
  }
  \label{fig:error}
\end{figure*}

\textbf{Error Loss without Parameter Tying.} 
To further investigate the effectiveness of parameter tying, we then analyze the error loss of different tasks during training.
As illustrated in Fig. \ref{fig:error}, we compare the error loss by eliminating the parameter tying from Task-1 to Task-5 during different training epochs.
Our FedSTIL with the parameter tying can achieve lower loss and converge much faster than eliminating the parameter tying. 
The improvement is mainly because parameter tying enables on-edge models to continuously fit new tasks with minimal parameter changes and forgetting of previous knowledge learnt in the past.
Thus, the design of tying the spatial-temporal correlated edge models for jointly optimizing can achieve better convergence and generalization.


\begin{table}[t]
  \caption{Performances under different distance metrics.}
  \label{tab:distance}
  \centering
  \setlength{\tabcolsep}{2mm}{
  \begin{tabular}{ccccc}
    \toprule
    Distance & mAP \small{(\%)} & R1 \small{(\%)} & R3 \small{(\%)} & R5 \small{(\%)} \\
    \midrule
    Cosine & 66.92 & 65.13 & 70.92 & 72.92 \\
    Euclidean & 67.03 & 65.27 & 70.88 & 72.93 \\
    KL & 68.16 & 66.05 & 72.03 & 74.05 \\
  \bottomrule
\end{tabular}}
\end{table}

\textbf{Performances under Different Distance Metrics.} 
Considering the characteristics of the task features, we adopt Kullback-Leibler (KL) Divergence to compute the spatial-temporal correlation ratio when aggregating the personalized sharing knowledge. 
We also evaluate the retrieval accuracy of our method under some other distance metrics, namely, Cosine Distance and Euclidean Distance. 
%
As shown in Table \ref{tab:distance}, KL can achieve higher accuracy than other metrics. 
Different from Cosine and Euclidean, KL can effectively measure the information difference among the different distributions of task features.
Hence, KL can help the parameter server precisely measure the edge similarities to aggregate the spatial-temporal knowledge with low noise and less interference.

\begin{figure}[h]
  \centering
  \includegraphics[width=\linewidth]{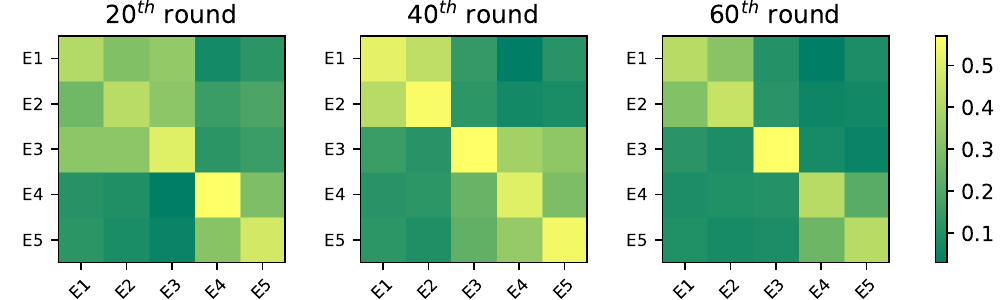}
  \caption{
  The visualization of the correlation weights at 20$^{th}$, 40$^{th}$ and 60$^{th}$ rounds. X-axis and Y-axis show the edge IDs.
  }
  \label{fig:round_vis}
\end{figure}

\begin{figure*}[h]
  \centering
  \includegraphics[width=\linewidth]{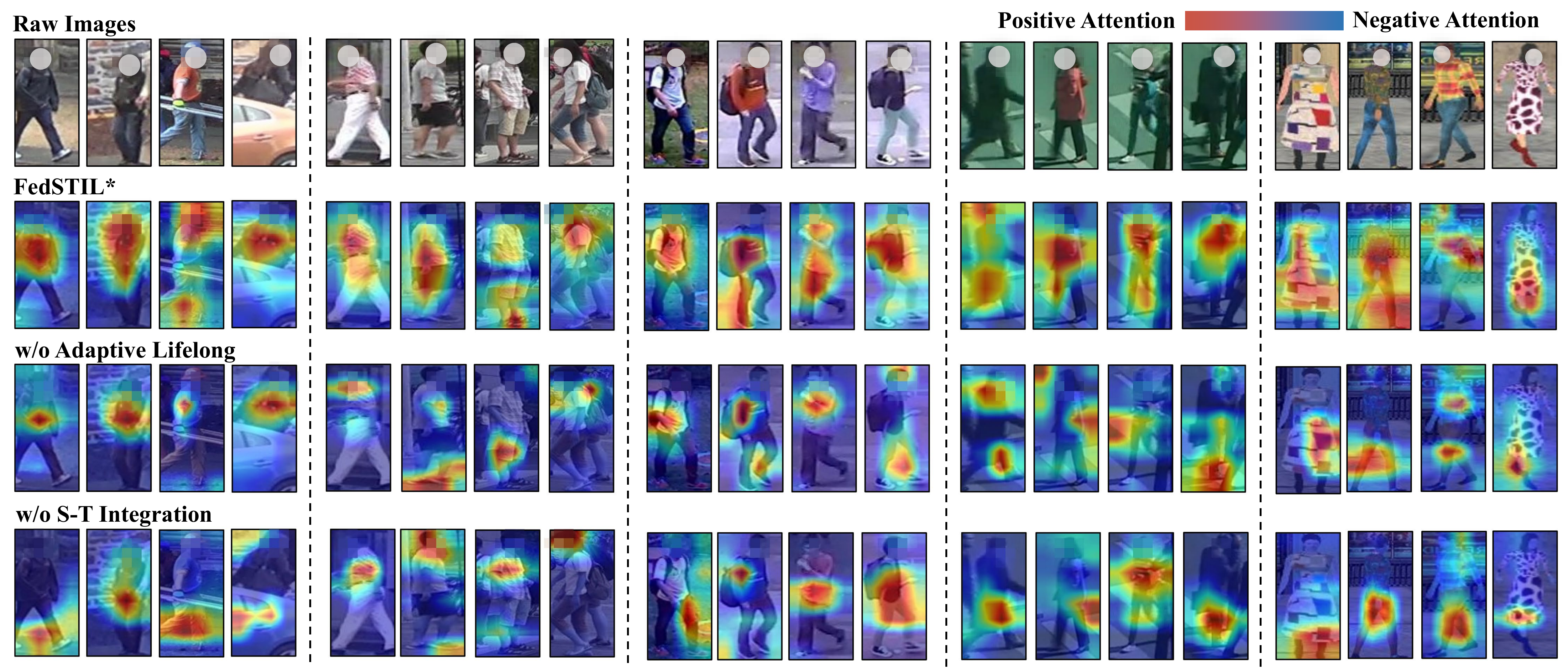}
  \caption{
  The heat maps of sampled person images from five datasets.
  The hot-spots are what the models focus on, which also reflect the generalization degree of representation.
  Images with red and blue boundaries denote positive attention and negative attention from the perspective of models.
  }
  \label{fig:visualization}
\end{figure*}

\textbf{Edge Correlations at Different Rounds.} 
To better understand the effectiveness of our spatial-temporal knowledge integration, we visualize the edge task correlation weights $W_{ij}^{(t)}$ calculated by our parameter server for all edges at 20$^{th}$, 40$^{th}$ and 60$^{th}$ round as shown in Fig. \ref{fig:round_vis}.
Specifically, we pre-define the ground-truth correlation groups that the 20$^{th}$ round are 1-3, 4-5; the 40$^{th}$ round are 1-2, 3-5; the 60$^{th}$ round are 1-2, 3, 4-5.
As shown in Fig. \ref{fig:round_vis}, we can observe that the pair-wise correlation weights with light color almost corresponds to the ground-truth, which reflect the effectiveness of our spatial-temporal knowledge integration.

\subsection{Visualization Study}
To further verify the effectiveness of some designs in our method FedSTIL, we visualize the heat-maps of our method by removing adaptive lifelong learning and spatial-temporal integration to demonstrate their influences on the representation.
We utilize the gradient-weighted class activation map (GradCAM)  \cite{selvaraju2017grad} to generate the heat maps, where the hot-spots are the positive focus of the models for re-identification.

We can observe in Fig. \ref{fig:visualization} that without the spatial-temporal knowledge integration, the model will only focus on some limited regions, such as shoes, coats, or bags. 
On the contrary, the attention regions of our oracle method FedSTIL are mainly on the overall person's body and have more generalized representations.
This improvement is largely due to the effectiveness of knowledge sharing to improve the generalizing capacities for edge clients. 
Moreover, if without our adaptive lifelong learning, we can also observe that the model attention may deviate from the regions of person body, and may even focus on the surrounding or background regions. 
This is probably because the learnt knowledge from the prior tasks will be gradually forgotten during the lifelong learning of new scenarios. 
However, our FedSTIL with the prototype rehearsal will mitigate the forgetting of the former knowledge and maintain the focus on person regions during lifelong learning.

\section{Future Work} \label{sec:future_work}

\textbf{Domain Generalization.}
The edge clients may have completely different source domains.
To adapt to diverse domains, distributed person ReID models may require frequent retraining or fine-tuning, resulting in large training time and resource costs.
To better adapt to unknown domains in the open-world scenario, we will consider to implement domain generalization techniques \cite{yang2022federated, yang2021joint, shu2015weakly, zhao2020deep} to improve model generalization in the future.

\textbf{Heterogeneous Backbones.}
The distributed person ReID system may consist of heterogeneous edge devices with different computing capacity \cite{bhardwaj2022ekya}.
For instance, some edge devices with GPU accelerators (e.g., Jetson TX2) can support the lifelong learning with ResNet50,
while some weak devices (e.g., Raspberry Pi 4B) can barely afford MobileNetV2.
To meet the diverse computing capacities, the edge clients need to be deployed with different backbone networks.
Thus, we will consider how to support knowledge sharing under heterogeneous backbones among distributed edges in the future.

\textbf{Occluded Person ReID.}
The real-world person ReID scenarios are occasionally crowded, and the captured person images may be overlapped or occluded by some obstacles.
Occluded person ReID significantly differs from the conventional person ReID due to the lack of full-body person images.
However, the missing information of the occluded person images may be complemented by judicious organizing spatial-temporal knowledge among distributed edges.
Therefore, implementing the occluded person ReID techniques \cite{dou2022human, zhao2021incremental} into our framework to enhance model robustness is also a promising direction for future work.

\section{Conclusion} \label{sec:conclusion}
In this work, we proposed a federated lifelong learning framework, FedSTIL, which enables the distributed edge clients to learn collaboratively and continuously in real-life for person ReID in new scenarios.
Our proposed method can utilize the spatial-temporal correlated knowledge among the edge clients to improve the data representation for person ReID and achieve better performance.
Meanwhile, the learnt knowledge from the historical tasks can remain long-term effective with only slight forgetting during continuously learning for new scenarios.
Experimental results verified that FedSTIL improves the accuracy of person  ReID while reducing the communication cost for distributed learning. 

\section*{Acknowledgment} \label{sec:ack}
This research was supported by the Natural Science Foundation of Jiangsu Province under Grant BK20200486 and NSFC No. 61972086.

\bibliographystyle{IEEEtran}
\bibliography{flreid_paper}{}

\newpage

\vfill

\end{document}